\documentclass{article}
\usepackage[labelfont=bf]{caption}
\usepackage{titlesec}
\usepackage[margin=1.5in]{geometry}

\usepackage[nottoc,notlof,notlot]{tocbibind} 
\usepackage[nottoc]{tocbibind}
\usepackage{indentfirst}
\usepackage[utf8]{inputenc} 
\usepackage[T1]{fontenc}    
\usepackage{hyperref}       
\usepackage{url}            
\usepackage{booktabs}       
\usepackage{amsfonts}       
\usepackage{nicefrac}       
\usepackage{microtype}      
\usepackage{graphicx}
\usepackage{amssymb}
\usepackage[utf8]{inputenc}
\usepackage{lipsum}
\usepackage[english]{babel}
\usepackage{hyperref}
\usepackage{url}
\usepackage{booktabs}
\usepackage{verbatim}
\usepackage{setspace}
\usepackage{graphicx}
\usepackage[centertags]{amsmath}
\usepackage{amsfonts}
\usepackage{multirow}
\usepackage{multicol}
\usepackage{amssymb}
\usepackage{amsmath}
\usepackage{amsthm}
\usepackage[noend]{algpseudocode}
\usepackage{algorithm}

\let\emptyset\varnothing
\usepackage{enumitem}
\setlist[enumerate]{itemsep=0mm}

\def\utilde#1{\mathord{\vtop{\ialign{##\crcr
$\hfil\displaystyle{#1}\hfil$\crcr\noalign{\kern1.5pt\nointerlineskip}
$\hfil\tilde{}\hfil$\crcr\noalign{\kern1.5pt}}}}}
\def\balign#1\ealign{\begin{align}#1\end{align}}
\def\baligns#1\ealigns{\begin{align*}#1\end{align*}}
\def\bitemize#1\eitemize{\begin{itemize}#1\end{itemize}}
\def\benumerate#1\eenumerate{\begin{enumerate}#1\end{enumerate}}
 
\def\reals{\mathbb{R}} 

\def\P{\mathcal{P}}
 
\def\Bet{\textnormal{beta}}

\newtheorem{theorem}{Theorem}
\newtheorem{corollary}[theorem]{Corollary}
\newtheorem{lemma}[theorem]{Lemma}

\newtheorem{remark}{Remark}[section]
\newtheorem{definition}[theorem]{Definition}

\title{Density Estimation via Discrepancy Based \\ Adaptive  Sequential Partition}

\author{
Dangna Li\\
ICME\\
Stanford University\\
Stanford, CA 94305\\
\texttt{dangna@stanford.edu} \\
\and
Kun Yang \\
Google\\
Mountain View, CA 94043\\
\texttt{kunyang0830@gmail.com} \\
\and
Wing Hung Wong\\
Department of Statistics \\
Stanford University \\
 Stanford, CA 94305\\
\texttt{whwong@stanford.edu} 
}
\date{}
\begin{document}

\maketitle

\begin{abstract}
Given $iid$ observations from an unknown absolute continuous distribution defined on some domain $\Omega$, we propose a nonparametric method to learn a piecewise constant function to approximate the underlying probability density function. Our density estimate is a piecewise constant function defined on a binary partition of $\Omega$.  The key ingredient of the algorithm is to use discrepancy, a concept originates from Quasi Monte Carlo analysis, to control the partition process. The resulting algorithm is simple, efficient, and has a provable convergence rate. We empirically demonstrate its efficiency as a density estimation method. We present its applications on a wide range of tasks, including finding good initializations for k-means. 

\end{abstract}
\section{Introduction}
Density estimation is one of the fundamental problems in statistics. Once an explicit estimate of the density function is constructed, various kinds of statistical inference tasks follow naturally. Given $iid$ observations, our goal in this paper is to construct an estimate of their common density function via a nonparametric domain partition approach.

As pointed out in   \cite{Lu2013}, for density estimation, the bias due to the limited approximation power of a parametric family will become dominant in the over all error as the sample size grows. Hence it is necessary to adopt a nonparametric approach to handle this bias. The kernel density estimation   \cite{parzen1962estimation} is a popular
nonparametric density estimation method. Although in theory it can achieve optimal
convergence rate when the kernel and the bandwidth are
appropriately chosen, its result can be sensitive to the choice of bandwidth, especially in high dimension. In practice, kernel density estimation is typically not applicable to problems of dimension higher than 6. 

Another widely used nonparametric density estimation method in low dimension is the histogram. But similarly with kernel density estimation, it can not be scaled easily to higher dimensions. Motivated by the usefulness of histogram and the need for a method to handle higher dimensional cases, we propose a novel nonparametric density estimation method which learns a piecewise constant density function defined on a binary partition of domain $\Omega$.

A key ingredient for any partition based method is the decision for stopping. Based on the observation that for any piecewise constant density, the distribution conditioned on each sub-region is uniform, we propose to use star discrepancy, which originates from analysis of Quasi-Monte Carlo methods, to formally measure the degree of uniformity. We will see in  section $\ref{theory}$ that this allows our density estimator to have near optimal convergence rate. 

In summary, we highlight our contribution as follows:
\begin{itemize}[itemsep=-0.5mm]
	\item To the best of our knowledge, our method is the first density estimation method that utilizes Quasi-Monte Carlo technique in density estimation.
	\item 
	We provide an error analysis on binary partition based density estimation method. We establish an $O(n^{-\frac{1}{2}})$ error bound for the density estimator. The result is optimal in the sense that essentially all Monte Carlo methods have the same convergence rate. Our simulation results support the tightness of this bound.
	\item One of the advantage of our method over existing ones is its efficiency.
 We demonstrate in section $\ref{sec:exp}$ that our method has comparable accuracy with other methods in terms of Hellinger distance while achieving an approximately $10^2$-fold speed up. 
		\item  Our method is a general data exploration tool and is readily applicable to many important learning tasks. Specifically, we demonstrate in section $\ref{sec:k-means}$ how it can be used to find good initializations for k-means.
\end{itemize}

\section{Related work}

Existing domain partition based density estimators can be divided into two categories: the first category belongs to the Bayesian nonparametric framework. Optional P\'{o}lya Tree (OPT)  \cite{Wong2010} is a class of nonparametric conjugate priors on the set of piecewise constant density functions defined on some partition of $\Omega$. Bayesian Sequential Partitioning (BSP)  \cite{Lu2013} is introduced as a computationally more attractive alternative to OPT. Inferences for both methods are performed by sampling from the posterior distribution of density functions. Our improvement over these two methods is two-fold. First, we no longer restrict the binary partition to be always at the middle. By introducing a new statistic called the ``gap'', we allow the partitions to be adaptive to the data. Second,  our method does not stem from a Bayesian origin and proceeds in a top down, greedy fashion. This makes our method computationally much more attractive than OPT and BSP, whose inference can be quite computationally intensive. 

The second category is tree based density estimators  \cite{liu2011forest} \cite{ram2011density}. As an example, Density Estimation Trees  \cite{ram2011density} is generalization of classification trees and regression trees for the task of density estimation. Its tree based origin has led to a loss minimization perspective: the learning of the tree is done by minimizing the integrated squared error. However, the true loss function can only be approximated by a surrogate and the optimization problem is difficult to solve. The objective of our method is much simpler and leads to an intuitive and efficient algorithm.

\section{Main algorithm}\label{method}
\subsection{Notations and definitions} 
In this paper we consider the problem of estimating a joint density function $f$ from a given set of observations. Without loss of generality, we assume the data domain $\Omega = [0,1]^d$, a hyper-rectangle in $\reals{^d}$. We use the short hand notation $[a,b]=\prod_{j=1}^d[a_j,b_j]$ to denote a hyper-rectangle in $\reals{^d}$, where $a=(a_1,\cdots,a_d), b=(b_1,\cdots,b_d)\in[0,1]^d$. Each $(a_j,b_j)$ pair specifies the lower and upper bound of the hyper-rectangle along dimension $j$.  

We restrict our attention to the class of piecewise constant functions after balancing the trade-off between simplicity and representational power: Ideally, we would like the function class to have concise representation while at the same time allowing for efficient evaluation. On the other hand, we would like to be able to approximate any continuous density function arbitrarily well (at least as the sample size goes to infinity). This trade-off has led us to choose the set of piecewise constant functions supported on binary partitions: First, we only need $2d+1$ floating point numbers to uniquely define a sub-rectangle ($2d$ for its location and $1$ for its density value). Second, it is well known that the set of positive, integrable, piesewise constant functions is dense in $L^p$ for $p\in[1,\infty)$.

\begin{figure}
 \center
 \includegraphics[width = .9\textwidth]{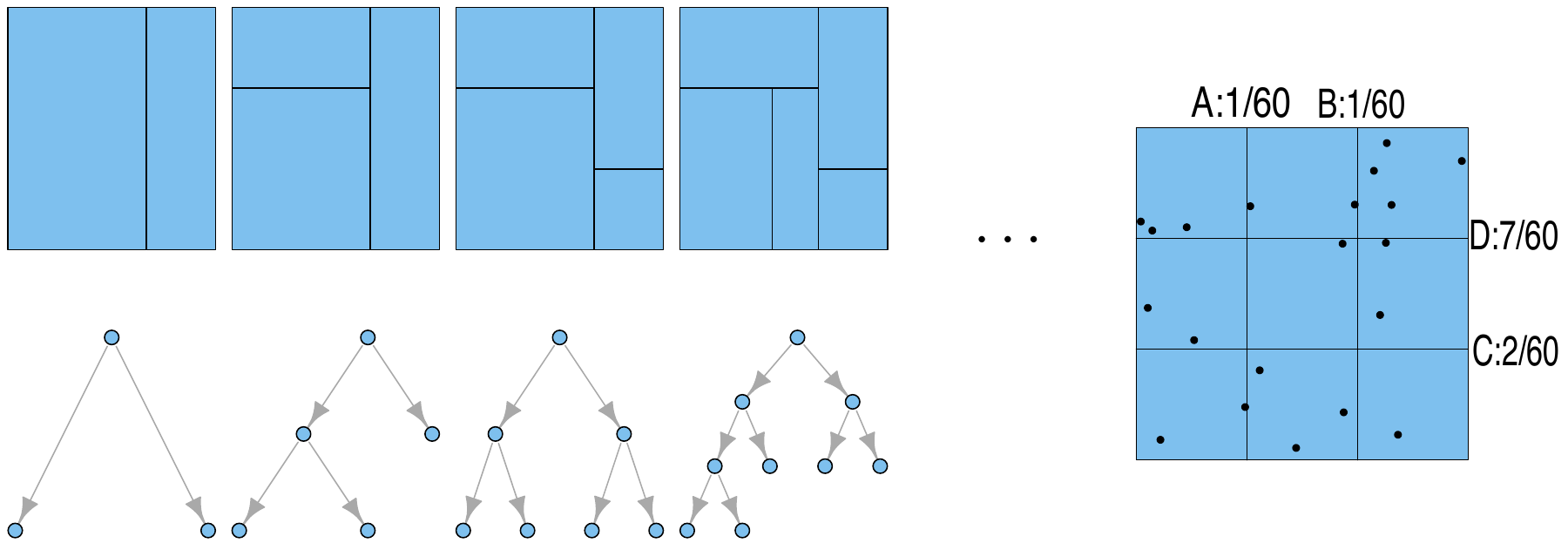}
 \caption{Left: a sequence of binary partition and the corresponding tree representation; if we encode partitioning information (e.g., the location where the split occurs) in the nodes, there is a one to one mapping between the tree representations and the partitions. Right: the gaps with $m = 3$, we split the rectangle at location D, which corresponds to the largest gap (Assuming it does not satisfy \eqref{cond}, see the text for more details)}.
  \label{fig:bp}
\end{figure}

The binary partition we consider can be defined in the following recursive way: starting with $\P_0 = {\Omega}$. Suppose we have a binary partition $\P_t = \{\Omega^{(1)},\cdots,\Omega^{(t)}\}$ at level $t$, where $\cup_{i=1}^t \Omega^{(i)}=\Omega$, $\Omega^{(i)}\cap \Omega^{(j)}=\emptyset$, $i\neq j$, a level $t+1$ partition $\P_{t+1}$ is obtained by dividing one sub-rectangle $\Omega^{(i)}$ in $\P_t$ along one of its coordinates, parallel to one of the dimension. See Figure $\ref{fig:bp}$ for an illustration.


\subsection{Adaptive partition and discrepancy control}
The above recursive build up has two key steps. The first is to decide whether to further split a sub-rectangle. One helpful intuition is that for piecewise constant densities, the distribution conditioned on each sub-rectangle is uniform. Therefore the partition should stop when the points inside a sub-rectangle are approximately uniformly scattered. In other words, we stop the partition when further partitioning does not reveal much additional information about the underlying density landscape. We propose to use star discrepancy, which is a concept originates from the analysis of Quasi-Monte Carlo methods, to formally measure the degree of uniformity of points in a sub-rectangle. Star discrepancy is defined as:

\begin{definition}
    Given $n$ points $X_n = \{x_1, ..., x_n\} $ in $ [0, 1]^d$. The star discrepancy $D^*(X_n)$ is defined as:
    \small
    \begin{equation}
    D^*(X_n) = \sup_{a\in[0, 1]^d}\Big|\frac{1}{n}\sum_{i = 1}^n\mathbf{1}\{x_i\in[0, a)\} - \prod_{j = 1}^da_j\Big|
    \label{DI}
    \end{equation}
    \normalsize
\end{definition}

The supremum is taken over all $d$-dimensional sub-rectangles $[0,a)$.
Given star discrepancy $D^*(X_n)$, we have the following error bound for Monte Carlo integration (See  \cite{Kuipers2012} for a proof):
\begin{theorem}\label{KH}
    (Koksma-Hlawka inequality) Let $X_n = \{x_1, x_2, ..., x_n\}$ be a set of points in $[0, 1]^d$ with discrepancy $D^*(X_n)$; Let $f$ be a function on $[0, 1]^d$ of bounded variation $\mathcal{V}(f)$. Then,
    \small
    \[\Big|\int_{[0, 1]^d}f(x)dx - \frac{1}{n}\sum_{i = 1}^n f(x_i)\Big|\leq \mathcal{V}(f)D^*(X_n)\]
    \normalsize
    where $\mathcal{V}(f)$ is the total variation in the sense of Hardy and Krause (See  \cite{Owen2005} for its precise definition).
 \end{theorem}
The above theorem implies if the star discrepancy $D^*(X_n)$ is under control, the empirical distribution will be a good approximation to the true distribution. Therefore, we may decide to keep partitioning a sub-rectangle until its discrepancy is lower than some threshold. We shall see in section $\ref{theory}$ that this provably guarantees our density estimate is a good approximation to the true density function. 

Another important ingredient of all partition based methods is the choice of splitting point. In order to find a good location to split for $[a, b] = \prod_{j = 1}^d[a_j, b_j]$,  we divide $j^{th}$ dimension into $m$ equal-sized bins: $[a_j, a_j + (b_j - a_j) / m], ..., [a_j + (b_j - a_j) (m - 2) / m, a_j + (b_j - a_j) (m - 1) / m]$ and keep track of the gaps at $a_j + (b_j - a_j) / m, ..., a_j + (b_j - a_j) (m - 1) / m$, where the gap $g_{jk}$ is defined as $|(1/n)\sum_{i = 1}^n\mathbf{1}(x_{ij} < a_j + (b_j - a_j) k / m) - k / m|$ for $k = 1, ..., (m - 1)$, there are total $(m - 1)d$ gaps recorded (Figure \ref{fig:bp}).  Here $m$ is a hyper-parameter chosen by the user. $[a, b]$ is split into two sub-rectangles along the dimension and location corresponding to maximum gap (Figure \ref{fig:bp}). The pseudocode for the complete algorithm is given in Algorithm \ref{algo}. We refer to this algorithm as DSP in the sequel. One distinct feature of DSP is it only requires the user to specify two parameters: $m,\theta$, where $m$ is the number of bins along each dimension; $\theta$ is the parameter for discrepancy control (See theorem \ref{cond} for more details). In some applications, the user may prefer putting an upper bound on the number of total partitions. In that case, there is typically no need to specify $\theta$. Choices for these parameters are discussed in Section $\ref{sec:exp}$.


The resulting density estimates $\hat{p}$ is a piecewise constant function defined on a binary partition of $\Omega$: $
\hat{p}(x) = \sum_{i = 1}^L d(r_i)\mathbf{1}\{x\in r_i\}
 \label{eq1}
$ where $\mathbf{1}$ is the indicator function; $L$ is the total number of sub-rectangles in the final partition; $\{r_i, d(r_i)\}_{i = 1}^L$ are the sub-rectangle and density pairs.  We demonstrate in section $\ref{sec:exp}$ how $\hat{p}(x)$ can be leveraged to find good initializations for k-means. In the following section, we establish a convergence result of our density estimator.

\begin{algorithm}

\caption{Density Estimation via Discrepancy Based Sequential Partition (DSP)}
\textbf{Input}: $X_N, m, \theta$\\
\textbf{Output}: A piecewise constant function $\Pr(\cdot)$ defined on a binary partition $\mathcal{R}$\\
Let $\Pr(r)$ denote the probability mass of region $r\subset \Omega$; let $X_N(r)$ denote the points in $X_N$ that lie within $r$, where $r\subset \Omega$. $n_i$ denotes the size of set $X^{(i)}$.
{\fontsize{8}{1}\selectfont
\begin{algorithmic}[1]

\Procedure{DSP}{$X_N, m, \theta$}
\State $\mathcal{B} = \{[0, 1]^d\}$, $\Pr([0, 1]^d) = 1$
\While{true}
\State $\mathcal{R}' = {\emptyset}$
\For{each $r_i = [a^{(i)}, b^{(i)}]$ in $\mathcal{R}$}
\State Calculate gaps $\{g_{jk}\}_{j = 1, ..., d, k = 1, ..., m - 1}$
\State Scale $X(r_i) = \{x_{i_l}\}_{l = 1}^{n_i}$ to $\tilde{X}^{(i)} = \{\tilde{x}_{i_l} = (\frac{x_{i_l, 1} - a^{(i)}_{1}}{b^{(i)}_{1}}, ..., \frac{x_{i_l, d} - a^{(i)}_{d}}{b^{(i)}_{d}})\}_{l = 1}^{n_i}$
\If{$X(r_i)\neq\emptyset$ and $D^*(\tilde{X}^{(i)}) > \theta\sqrt{N}/n_i$}\Comment Condition \eqref{cond} in Theorem \ref{theorem3}
\State Split $r_i$ into $r_{i_1} = [a^{(i_1)}, b^{(i_1)}]$ and $r_{i_2} = [a^{(i_2)}, b^{(i_2)}]$ along the max gap (Figure \ref{fig:bp}).
\State $\Pr(r_{i_1}) = \Pr(r_i)\frac{|P(r_{i_1})|}{n_i}$, $\Pr(r_{i_2}) = \Pr(r_i) - \Pr(r_{i_1})$
\State $\mathcal{R}' = \mathcal{R}'\cup\{r_{i_1}, r_{i_2}\}$
\Else $\ \mathcal{R}' = \mathcal{R}'\cup\{r_i\}$
\EndIf
\EndFor
\If{$\mathcal{R}'\neq\mathcal{R}$}
{$\mathcal{R} = \mathcal{R}'$}
\Else $\ 
$return $\mathcal{R}, \Pr(\cdot)$
\EndIf
\EndWhile
\EndProcedure
\end{algorithmic}
}
\label{algo}
\end{algorithm}

\section{Theoretical results} \label{theory}
Before we establish our main theorem, we need the following lemma:\footnote{ The proof for Lemma \ref{UB}  can be found in  \cite{Heinrich2000}.
Theorem \ref{theorem3} and Corollary $\ref{cor:tv}$ are proved in the supplementary material.}

  \begin{lemma}
    Let $D_n^* = \inf_{\{x_1, ..., x_n\}\in [0, 1]^d}D^*(x_1, ..., x_n)$, then we have \small\[D_n^* \leq c\sqrt{\frac{d}{n}}\]\normalsize for all $n, d \in\reals{^+}$, where $c$ is some positive constant.
    \label{UB}
  \end{lemma}

We now state our main theorem: 
\begin{theorem}
   Let $f$ be a function defined on $\Omega=[0,1]^d$ with bounded variation. Let $X_N = \{x_1, ..., x_N \in \Omega\}$ and $\{[a^{(i)},b^{(i)}], i=1,\cdots,L\}$ be a level $L$ binary partition of $\Omega$. Further denote by $X^{(i)} = \{x_j = (x_{j1}, ..., x_{jd}), x_j\in[a^{(i)},b^{(i)}] \text{ and }\}\cap X_N$, i.e. the part of $X_N$ in sub-rectangle $i$. $n_i= |X^{(i)}|$. Suppose in each sub-rectangle $[a^{(i)}, b^{(i)}]$, $X^{(i)}$ satisfies
    \small
    \begin{equation}
    D^*(\tilde{X}^{(i)})\leq \alpha^{(i)}{D}_{n_i}^*
    \label{cond}
    \end{equation}
    \normalsize
    where $\tilde{X}^{(i)} = \{\tilde{x}_j = (\frac{x_{j1} - a^{(i)}_{1}}{b^{(i)}_{1}}, ..., \frac{x_{jd} - a^{(i)}_{d}}{b^{(i)}_{d}}), x_j\in X^{(i)}\}$
, $\alpha^{(i)}= \sqrt{\frac{N}{n_id}}\frac{\theta}{c}$ for some positive constant $\theta$, ${D}_{n_i}^*$ is defined as in lemma \ref{UB}. Then
    \small
    \begin{equation}
    \Big|\int_{[0, 1]^d}f(x)\hat{p}(x)dx - \frac{1}{N}\sum_{i = 1}^Nf(x_i)\Big|\leq \frac{\theta}{\sqrt{N}} \mathcal{V}(f)
    \label{converge}
    \end{equation}

    where $\hat{p}(x)$ is a piecewise constant density estimator given by
    \small
    \[\hat{p}(x) = \sum_{i = 1}^Ld_i\mathbf{1}\{x\in [a^{(i)}, b^{(i)}]\}\]
    \normalsize
   with $d_i = (\prod_{j = 1}^d(b_{j}^{(i)} - a_{j}^{(i)}))^{-1}n_i / N$, i.e., the empirical density. \normalsize
    \label{theorem3}
\end{theorem}
In the above theorem, $\alpha^{(i)}$ controls the relative uniformity of the points and is adaptive to $X^{(i)}$. It imposes more restrictive constraints on regions containing larget proportion of the sample ($n_i / N$). Although our density estimate is not the only estimator which satisfies \eqref{converge}, (for example, both the empirical distribution in the asymptotic limit and  kernel density estimator with sufficiently small bandwidth meet the criterion), one advantage of our density estimator is that it provides a very concise summary of the data while at the same time capturing the landscape of the underlying distribution. In addition, the piecewise constant function does not suffer from having too many ``local bumps'', which is a common problem for kernel density estimator. Moreover, under certain regularity conditions (e.g. bounded second moments), the convergence rate of Monte Carlo methods for $\frac{1}{N}\sum_{i = 1}^Nf(x_i)$ to $\int_{[0, 1]^d}f(x)p(x)dx$ is of order $O(N^{-\frac{1}{2}})$. Our density estimate is optimal in the sense that it achieves the same rate of convergence. Given theorem \ref{theorem3}, we have the following convergence result:
  
   \begin{corollary}\label{cor:tv}
    Let $\hat{p}(x)$ be the estimated density function as in theorem \ref{theorem3}. For any hyper-rectangle $A = [a, b]\subset [0, 1]^d$, let $\hat{P}(A) = \int_{A}\hat{p}(x)dx$ and $P(A) = \int_{A}p(x)dx$, then $$\sup_{A\subset [0,1]^d}|\hat{P}(A) - P(A)|\rightarrow 0$$ at the order  $O(n^{-\frac{1}{2}})$.
  \end{corollary}
  \begin{remark}
    It is worth pointing out that the total variation distance between two probability measures $\hat{P}$ and $P$ is defined as
    $\delta(\hat{P}, P) = \sup_{A\in\mathcal{B}}|\hat{P}(A) - P(A)|$, where $\mathcal{B}$ is the Borel $\sigma$-algebra of $[0, 1]^d$. In contrast, Corollary \ref{cor:tv} restricts $A$ to be hyper-rectangles.
  \end{remark}


\section{Experimental results}\label{sec:exp}
\subsection{Implementation details}
In some applications, we find it helpful to first estimate the marginal densities for each
component variables $x_{. j}$ $(j = 1,...,d)$, then make a copula
transformation $z_{.j}$= $\hat{F}_
j (x_{. j} )$, where $\hat{F}_
j$ is the estimated cdf of
$x_{.j}$. After such a transformation, we
can take the domain to be $[0, 1]^d$. Also we find this can save the number of partition needed by DSP. Unless otherwise stated, we use copula transform in our experiments whenever the dimension exceeds $3$.

  We make the following observations to improve the efficiency of DSP: 1) First observe that  $\max_{j = 1, ..., d} D^*(\{x_{ij}\}_{i = 1}^n)\leq D^*(\{x_i\}_{i = 1}^n)$. Let $x_{(i)j}$ be the $i$th smallest element in $\{x_{ij}\}_{i = 1}^n$, then $D^*(\{x_{ij}\}_{i = 1}^n) = \frac{1}{2n} + \max_{i }|x_{(i)j} - \frac{2i - 1}{2n}|$  \cite{Doerr2013}, which has complexity $O(n\log n)$. Hence $\max_{j = 1, ..., d} D^*(\{x_{ij}\}_{i = 1}^n)$ can be used to compare against $\theta\sqrt{L}/n$ first before calculating $D^*(\{x_i\}_{i = 1}^n)$; 2) $\theta\sqrt{N}/n$ is large when $n$ is small, but $D^*(\{x_i\}_{i = 1}^n)$ is bounded above by 1; 3) $\theta\sqrt{N}/n$ is tiny when $n$ is large and $D^*(\{x_i\}_{i = 1}^n)$ is bounded below by $c_d\log^{(d - 1)/2}n^{-1}$ with some constant $c_d$ depending on $d$  \cite{Gnewuch2012a}; thus we can keep splitting without checking \eqref{cond} when $\theta\sqrt{N}/n\leq\epsilon$, where $\epsilon$ is a small positive constant (say 0.001) specified by the user. This strategy has proved to be effective in decreasing the runtime significantly at the cost of introducing a few more sub-rectangles.

 Another approximation works well in practice is by replacing star discrepancy with computationally attractive $\mathcal{L}_2$ star discrepancy, i.e., $D^{(2)}(X_n) = (\int_{[0, 1]^d}|\frac{1}{n}\sum_{i = 1}^n\mathbf{1}_{x_i\in[0, a)} - \prod_{i = 1}^da_{i}|^2da)^{\frac{1}{2}}$; in fact, several statistics to test uniformity hypothesis based on $D^{(2)}$ are proposed in  \cite{Liang2001}; however, the theoretical guarantee in Theorem \ref{theorem3} no longer holds. By Warnock's formula  \cite{Doerr2013},
  \small
  \[[D^{(2)}(X_n )]^2 = \frac{1}{3^d} - \frac{2^{1 - d}}{n}\sum_{i = 1}^n\prod_{j = 1}^d(1 - x_{ij}^2) + \frac{1}{n^2}\sum_{i, l = 1}^n\prod_{j = 1}^d\min\{1 - x_{ij}, 1 - x_{lj}\}\]
  \normalsize
  $D^{(2)}$ can be computed in $O(n\log^{d - 1}n)$ by K. Frank and S. Heinrich's algorithm  \cite{Doerr2013}. At each scan of $\mathcal{R}$ in Algorithm \ref{algo}, the total complexity is at most $\sum_{i = 1}^L O(n_i\log^{d - 1}n_i)\leq\sum_{i = 1}^L O(n_i\log^{d - 1}N)\leq O(N\log^{d - 1}N)$.

There are no closed form formulas for calculating $D^*(X_n)$ and $D_{n}^*$ except for low dimensions. If we replace $\alpha^{(i)}$ in \eqref{cond} and apply Lemma \ref{UB}, what we are actually trying to do is to control $D^*(\tilde{X}^{(i)})$ by $\theta \sqrt{N}/n_i$. There are many existing work on ways to approximate $D^*(X_n)$. In particular, a new randomized algorithm based on threshold accepting is developed in  \cite{Gnewuch2012}. Comprehensive numerical tests indicate that it improves upon other algorithms, especially in when $20\leq d\leq 50$. We used this algorithm in our experiments. The interested readers are referred to the original paper for more details.
\subsection{DSP as a density estimate}\label{sec:simulation}
\begin{figure}[htbp]
    \center
    \includegraphics[width=.9 \textwidth]{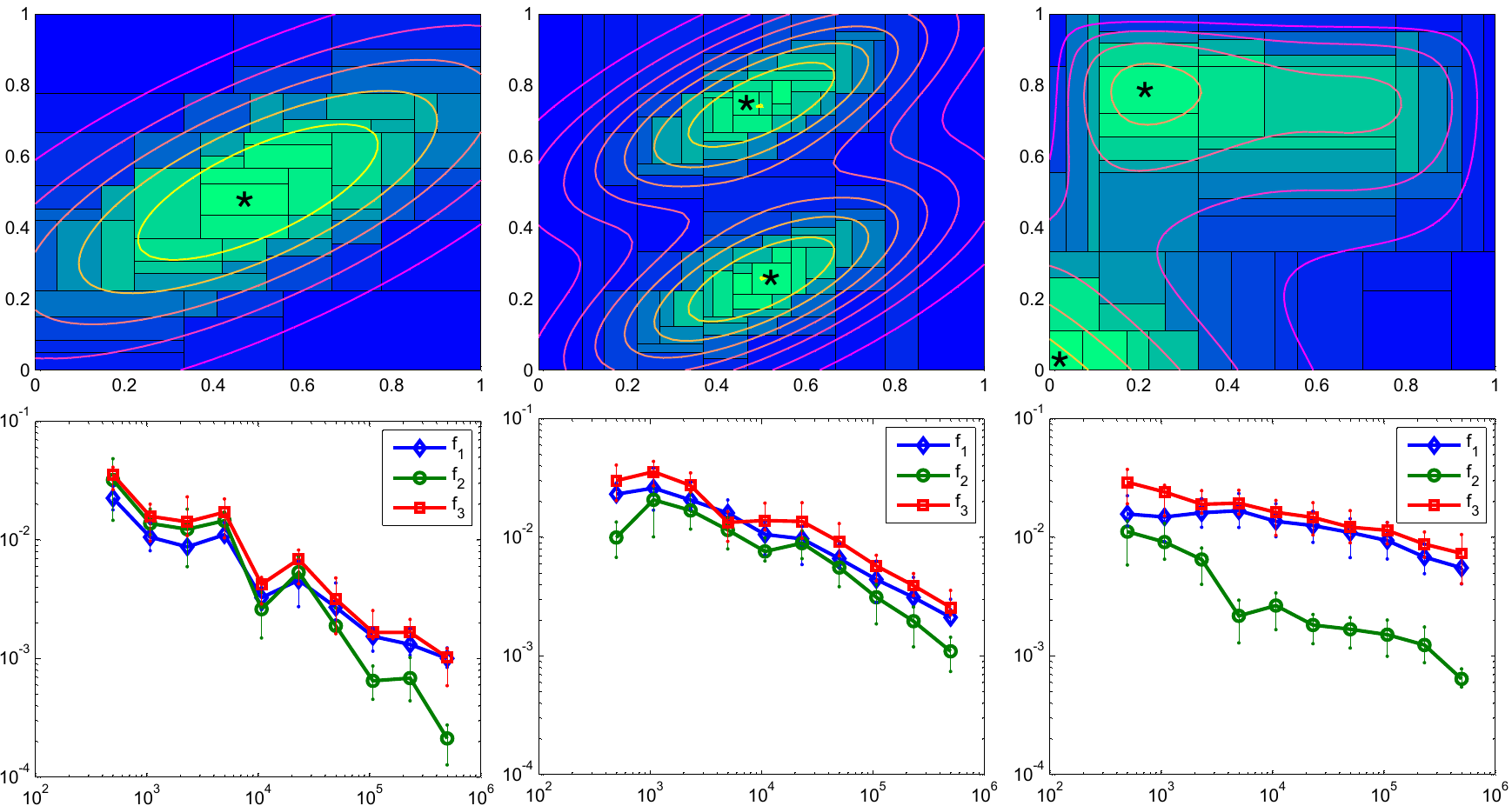}
    \caption{\textbf{First row}: estimated densities for 3 simulated 2D datasets. The modes are marked with stars.  The corresponding contours of true densities are embedded for comparison. \textbf{Second row}: simulation of 2, 5 and 10 dimensional cases (from left to right) with reference functions $f_1, f_2, f_3$. $x$-axis: sample size $n$. $y$-axis: error between the true integral and the estimated integral. The vertical bars are standard error bars obtained from 10 replications. See section \ref{sec:simulation} 2) for more details. }
    \label{simulation}
  \end{figure}

    \textbf{1)} To demonstrate the method and visualize the results, we apply it on several 2-dimensional data sets simulated from 3 distributions with different geometry:
        \begin{enumerate}[itemsep=0mm]
    \item Gaussian: $x\sim \mathcal{N}(\mu, \Sigma)\mathbf{1}\{x\in [0, 1]^2\},$ with $\mu=(.5,.5)^T$, $\Sigma = [0.08,0.02; 0.02, 0.02]$
    \item Mixture of Gaussians: $x\sim \frac{1}{2}\sum_{i=1}^2\mathcal{N}(\mu_i, \Sigma_{i})\mathbf{1}\{x\in [0, 1]^2\}$ with $\mu_1 = (.50, .25)^T,$ and $\mu_2 = (.50, .75)^T, \Sigma_1 = \Sigma_2 = [0.04, 0.01;0.01, 0.01]$;
  \item Mixture of Betas: $x\sim \frac{1}{3}(\Bet(2, 5)\Bet(5, 2) + \Bet(4, 2)\Bet(2, 4) + \Bet(1, 3)\Bet(3, 1))$;   
    \end{enumerate}
 where $\mathcal{N}(\mu,\Sigma)$ denotes multivariate Gaussian distribution and $\Bet(\alpha,\beta)$ denotes beta distribution. We simulated $10^5$ points for each distribution. See the first row of Figure \ref{simulation} for visualizations of the estimated densities. The figure shows DSP accurately estimates the true density landscape in these three toy examples.
 

    \textbf{2)} To evaluate the theoretical bound \eqref{converge}, we choose the following three 3 reference functions with dimension $d = 2$, 5 and 10 respectively: $f_1(x) = \sum_{i = 1}^n\sum_{j = 1}^dx_{ij}^{\frac{1}{2}}$, $f_2(x) = \sum_{i = 1}^n\sum_{j = 1}^dx_{ij}$,  $f_3(x) = (\sum_{i = 1}^n\sum_{j = 1}^dx_{ij}^{\frac{1}{2}})^2$. 
  We generate $n\in\{10^2, 10^3, 10^4,10^5,10^6\}$ samples from $p(x) = \frac{1}{2}\Big(\prod_{j = 1}^d\Bet(x_j, 15, 5) + \prod_{j = 1}^d\Bet(x_j, 5, 15)\Big)$, where $\Bet(\cdot,\alpha,\beta)$ is the density function of beta distribution.
  
  The error $|\int_{[0, 1]^d}f_k(x)p(x)dx - \int_{[0, 1]^d}f_k(x)\hat{p}(x)dx|$ is bounded by
  $|\int_{[0, 1]^d}f_k(x)p(x)dx - \frac{1}{n}\sum_{j = 1}^nf_k(x_j)| + |\int_{[0, 1]^d}f_k(x)\hat{p}(x)dx - \frac{1}{n}\sum_{j = 1}^n f_k(x_j)|$
  where $\hat{p}(x)$ is the estimated density; For  almost all Monte Carlo methods, the first term is of order $O(n^{-\frac{1}{2}})$. The second term is controlled by \eqref{converge}. Thus in total the error is of order $O(n^{-\frac{1}{2}})$. We have plot the error against the sample size on log-log scale for each dimension in the second row of Figure $\ref{simulation}$. The linear trends in the plots corroborate the bound in \eqref{converge}.

 \textbf{3)} To show the efficiency and scalability of DSP, we compare it with KDE, OPT and BSP in terms of estimation error and running time. We simulate samples from $x\sim (\sum_{i = 1}^4\pi_i \mathcal{N}(\mu_i, \Sigma_i))\mathbf{1}\{x\in [0, 1]^d\}$ with $d = \{2, 3, \cdots, 6\}$ and $N = \{10^3, 10^4, 10^5\}$ respectively. The estimation error measured in terms of Hellinger Distance is summarized in Table $\ref{tab:hd}$.  We set $m=10$, $\theta=0.01$ in our experiments. We found the resulting Hellinger distance to be quite robust as  $m$ ranges from $3$ to $20$ (equally spaced). The supplementary material includes the exact details about the parameters of the simulating distributions, estimation of Hellinger distance and other implementation details for the algorithms. The table shows DSP achieves comparable accuracy with the best of the other three methods. As mentioned at the beginning of this paper, one major advantage of DSP's is its speed. Table $\ref{tab:rt}$ shows our method achieves a significant speed up over all other three algorithms. 
   
  \begin{table*}[htbp]
  \centering
    \caption{\textbf{Error in Hellinger Distance} between the true density and KDE, OPT, BSP, our method for each $(d, n)$  pair.  The numbers in parentheses are standard errors from 20 replicas. The best of the four method is highlighted in bold. Note that the simulations, being based on mixtures of Gaussians, is unfavorable for methods based on domain partitions. }
  \label{tab:hd}
 \resizebox{\textwidth}{!}{
\begin{tabular}{llccccccccccc}
\hline
    & \multicolumn{4}{c}{Hellinger Distance ($n = 10^3$)}             & \multicolumn{4}{c}{Hellinger Distance ($n = 10^4$)}                    & \multicolumn{4}{c}{Hellinger Distance ($n = 10^5$)}             \\ \hline
d   & KDE             & OPT             & BSP      & DSP             & KDE             & OPT             & BSP             & DSP             & KDE             & OPT             & BSP      & DSP             \\
$2$ & 0.2331          & \textbf{0.2147} & 0.2533   & 0.2634          & 0.1104          & 0.0957          & 0.1222          & \textbf{0.0803} & \textbf{0.0305} & 0.0376          & 0.0345   & 0.0312          \\
    & (0.0421)        & (0.0172)        & (0.0163) & (0.0207)        & (0.0102)        & (0.0036)        & (0.0043)        & (0.0013)        & (0.0021)        & (0.0021)        & (0.0025) & (0.0027)        \\
$3$ & \textbf{0.2893} & 0.3279          & 0.2983   & 0.3072          & 0.2003          & 0.1722          & \textbf{0.1717} & 0.1721          & 0.1466          & 0.1117          & 0.1323   & \textbf{0.1020} \\
    & (0.0227)        & (0.0128)        & (0.0133) & (0.0265)        & (0.0199)        & (0.0028)        & (0.0083)        & (0.0073)        & (0.0047)        & (0.0008)        & (0.0009) & (0.004)         \\
$4$ & 0.3913          & \textbf{0.3839} & 0.3872   & 0.3895          & \textbf{0.2466} & 0.2726          & 0.2882          & 0.2955          & 0.1900          & 0.1880          & 0.2100   & \textbf{0.1827} \\
    & (0.0325)        & (0.0136)        & (0.0117) & (0.0191)        & (0.0113)        & (0.0031)        & (0.0047)        & (0.0065)        & (0.0057)        & (0.0006)        & (0.0006) & (0.0059)        \\
$5$ & 0.4522          & 0.4748          & 0.4435   & \textbf{0.4307} & 0.3599          & \textbf{0.3562} & 0.3987          & 0.3563          & \textbf{0.2817} & 0.2822          & 0.2916   & 0.2910          \\
    & (0.0317)        & (0.009)         & (0.0167) & (0.0302)        & (0.0199)        & (0.0025)        & (0.0022)        & (0.0031)        & (0.0088)        & (0.0005)        & (0.0003) & (0.0002)        \\
$6$ & 0.5511          & \textbf{0.5508} & 0.5515   & 0.5527          & 0.4833          & 0.4015          & 0.4093          & \textbf{0.3911} & 0.3697          & \textbf{0.3409} & 0.3693   & 0.3701          \\
    & (0.0318)        & (0.0307)        & (0.0354) & (0.0381)        & (0.0255)        & (0.0023)        & (0.0046)        & (0.0037)        & (0.0122)        & (0.0005)        & (0.0004) & (0.0002)        \\ \hline
\end{tabular}}
  \end{table*}
\begin{table}[htbp]
\centering
\caption{\textbf{Average CPU time in seconds} of KDE, OPT, BSP and our method for each $(d, n)$ pair. The numbers in parentheses are standard errors from 20 replicas. The speed-up is the fold speed-up computed as the ratio between the minimum run time of the other three methods and the run time of DSP. All methods are implemented in \text{C++}.  See the supplementary material for more details.}
\label{tab:rt}
\resizebox{\textwidth}{!}{
\begin{tabular}{lccccccccccccccc}
\hline
                   & \multicolumn{5}{c}{Running time ($n=10^3$)}       & \multicolumn{5}{c}{Running time ($n=10^4$)}       & \multicolumn{5}{c}{Running time ($n=10^5$)}         \\ \hline
d                  & KDE     & OPT     & BSP     & DSP     & speed-up & KDE     & OPT     & BSP     & DSP     & speed-up & KDE       & OPT     & BSP     & DSP     & speed-up \\
\multirow{2}{*}{2} & 2.445   & 9.484   & 0.833   & 0.020   & 41  & 21.903  & 31.561  & 1.445   & 0.033   & 43   & 230.179   & 44.561  & 7.750   & 0.242   & 33     \\
                   & (0.191) & (0.029) & (0.006) & (0.002) &          & (1.905) & (0.079) & (0.014) & (0.002) &          & (130.572) & (0.639) & (0.178) & (0.015) &          \\
\multirow{2}{*}{3} & 2.655   & 25.073  & 1.054   & 0.019   & 55   & 26.964  & 36.683  & 2.819   & 0.044   & 64  & 278.075   & 56.329  & 21.104  & 0.378   & 55   \\
                   & (0.085) & (0.056) & (0.010) & (0.002) &          & (1.089) & (0.076) & 0.036)  & (0.001) &          & (10.576)  & (0.911) & (0.576) & (0.011) &          \\
\multirow{2}{*}{4} & 3.540   & 32.112  & 1.314   & 0.019   & 69   & 37.141  & 39.219  & 5.861   & 0.049   & 119  & 347.501   & 67.366  & 53.620  & 0.485   & 108  \\
                   & (0.116) & (0.072) & (0.014) & (0.002) &          & (2.244) & (0.221) & (0.076) & (0.002) &          & (14.676)  & (3.018) & (2.917) & (0.018) &          \\
\multirow{2}{*}{5} & 4.107   & 37.599  & 1.713   & 0.020   & 85   & 45.580  & 44.520  & 12.220  & 0.078   & 157 & 412.828   & 77.776  & 115.869 & 0.706   & 110  \\
                   & (0.110) & (0.088) & (0.019) & (0.002) &          & (2.124) & (0.587) & (0.154) & (0.002) &          & (16.252)  & (2.215) & (6.872) & (0.051) &          \\
\multirow{2}{*}{6} & 4.986   & 41.565  & 2.749   & 0.020   & 137   & 53.291  & 43.032  & 21.696  & 0.127   & 170   & 519.298   & 81.023  & 218.999 & 0.896   & 90  \\
                   & (0.214) & (0.147) & (0.024) & (0.001) &          & (2.767) & (0.413) & (0.213)  & (0.004) &          & (29.276)  & (3.703) & (6.046) & (0.071) &          \\ \cline{2-16} 
\end{tabular}
}

\end{table}  
\subsection{DSP-kmeans}\label{sec:k-means}

In addition to being a competitive density estimator, we demonstrate in this section how DSP can be used to get good initializations for k-means. The resulting algorithm is referred to as DSP-kmeans. 

Recall that  given a fixed number of clusters $K$, the goal of k-means is to minimize the following objective function:
\begin{equation}\label{k-meansobj}
	J_K \stackrel{\Delta}{=} \sum_{k=1}^K\sum_{i\in C_k}\|x_i - m_k\|_2^2
\end{equation}

where $C_k$ denote the set of points in cluster $k$; $\{m_k\}_{k=1}^K$ denote the cluster means. The original k-means algorithms proceeds by alternating between assigning points to centers and recomputing the means. As a result, the final clustering is usually only a local optima and can be sensitive to the initializations. Finding a good initialization has attracted a lot of attention over the past decade and now there is a descent number existing methods, each with their own perspectives. Below we review a few representative types. 

One type of methods look for good initial centers sequentially. The idea is once the first center is picked, the second should be far away from the one that is already chosen. A similar argument applies to the rest of the centers.  \cite{arthur2007k} \cite{katsavounidis1994new} fall under this category. Several studies  \cite{fraley1998algorithms}
 \cite{redmond2007method} borrow ideas from hierarchical agglomerative
clustering (HAC) to look for good initializations. In our experiments we used the algorithm described in  \cite{fraley1998algorithms}. One essential ingredient of this type of algorithms is the inter cluster distance, which could be problem dependent. Last but not least, there is a class of methods that attempt to utilize the relationship between PCA and k-means.  \cite{xu2015pca} proposes a PCA-guided search for initial centers. 
 \cite{su2007search} combines the relationship between PCA and k-means to look for good initialization. The general idea is to recursively
splitting a cluster according the first principal
component. We refer to this algorithm as PCA-REC.

DSP-kmeans is different from previous methods in that it tackles the initialization problem from a density estimation point of view. The idea behind DSP-kmeans is that cluster centers should be close to the modes of underlying probability density function. If a density estimator can accurately locate the modes of the underlying true density function, it should also be able to find good cluster centers.  Due to its concise representation, DSP can be used for finding initializations for k-means in the following way: Suppose we are trying to cluster a dataset $Y$ with $K$ clusters. We first apply DSP on $Y$ to find a partition with $K$ non-empty sub-rectangles, i.e. sub-rectangles that have at least one point from $Y$. The output of DSP will be $K$ sub-rectangles. Denote the set of indices for the points in sub-rectangle $j$ by $S_j$, $j = 1, \dots, K$, let $I_j = \frac{1}{|S_j|} \sum_{i\in S_j} Y_i$, i.e. $I_j $ is the sample average of points fall into sub-rectangle $j$. We then use $\{I_1,\cdots,I_K\}$ to initialize k-means. We also explored the following two-phase procedure: first over partition the space to build a more accurate density estimate. Points in different sub-rectangles are considered to be in different clusters. Then we merge the sub-rectangles hierarchically based on some measure of between cluster distance. We have found this to be helpful when the number of clusters $K$ is relatively small. For completeness, we have included the details of this two-phase DSP-kmeans in the supplementary material.

We test DSP-kmeans on 4 real world datasets of various number of data points and dimensions. Two of them are taken from the UCI machine learning repository  \cite{kaul2013building}; the stem cell data set is taken from the FlowCAP challenges  \cite{aghaeepour2013critical}; the mouse bone marrow data set is a recently published single-cell dataset measured using mass cytometry  \cite{spitzer2015interactive}. We use random initialization as the base case and compare it with DSP-kmeans, k-means++, PCA-REC and HAC. The numbers in Table \ref{tab:k-means} are the improvements in k-means objective function of a method over random initialization. The result shows when the number of clusters is relatively large DSP-kmeans achieves lower objective value in these four datasets. 
\begin{table}[htbp]
\centering
\caption{\textbf{Comparison of different initialization methods}. The number for method $j$ is relative to random initialization: $\frac{J_{K,j}- J_{K,0}}{J_{K,0}}$, where $J_{K,j}$ is the k-means objective value of method $j$ at convergence. Here we use $0$ as index for random initialization. Negative number means the method perform worse than random initialization. } \label{tab:k-means}
 \resizebox{\textwidth}{!}{
\begin{tabular}{llcccccccccccc}
\hline
             &                   &    & \multicolumn{4}{c}{Improvement over random init.}        &                       &                             &    & \multicolumn{4}{c}{Improvement over random init.}          \\ \hline
\multicolumn{2}{l}{Road network} & k  & k-means++     & PCA-REC & HAC           & DSP-kmeans     & \multicolumn{2}{l}{Mouse bone marrow}               & k  & k-means++      & PCA-REC & HAC            & DSP-kmeans     \\
n            & 4.3e+04           & 4  & 0.0             & -0.02   & \textbf{0.01} & 0.0              & \multicolumn{1}{l}{n} & \multicolumn{1}{l}{8.7e+04} & 4  & \textbf{1.51}  & 0.03    & 1.25           & 0.4            \\
d            & 3                 & 10 & 0.0             & -0.12   & \textbf{0.25} & 0.08           & \multicolumn{1}{l}{d} & \multicolumn{1}{l}{39}      & 10 & 0.45           & 0.24    & 0.77           & \textbf{0.83}  \\
             &                   & 20 & 0.43          & -0.46   & 1.68          & 2.04           &                       &                             & 20 & 0.63           & -1.2    & 0.68           & \textbf{0.79}  \\
             &                   & 40 & 11.7          & -2.52   & 2.27          & 13.62          &                       &                             & 40 & 1.99           & -3.56   & 2.06           & \textbf{2.55}  \\
             &                   & 60 & 19.78         & -3.45   & 18.69         & \textbf{20.91} & \multicolumn{2}{c}{}                                & 60 & 2.48           & -5.25   & 2.57           & \textbf{2.65}  \\ \hline
\multicolumn{2}{l}{Stem cell}    & k  & k-means++     & PCA-REC & HAC           & DSP-kmeans     & \multicolumn{2}{l}{US census}                       & k  & k-means++      & PCA-REC & HAC            & DSP-kmeans     \\
n            & 9.9e+03           & 4  & 3.45          & -2.1    & 3.67          & \textbf{3.96}  & \multicolumn{1}{l}{n} & \multicolumn{1}{l}{2.4e+06} & 4  & \textbf{47.44} & -2.33   & 46.72          & 40.44          \\
d            & 6                 & 10 & \textbf{3.82} & -4.2    & 3.79          & 3.6            & \multicolumn{1}{l}{d} & \multicolumn{1}{l}{68}      & 10 & 40.52          & -1.9    & \textbf{41.48} & 39.52          \\
             &                   & 20 & \textbf{9.96} & -3.59   & 9.91          & 9.39           &                       &                             & 20 & \textbf{32.63} & -1.97   & 29.49          & 32.55          \\
             &                   & 40 & 9.95          & -6.39   & 10.11         & \textbf{12.49} &                       &                             & 40 & 32.66          & -5.15   & 33.41          & \textbf{34.61} \\
             &                   & 60 & 6.12          & -7.29   & 8.19          & \textbf{13.7}  &                       &                             & 60 & \textbf{21.7}  & -1.19   & 16.28          & 21.68          \\ \hline
\end{tabular}
}
\end{table} 
Although in theory almost all density estimator could be used to find good initializations. Based on the comparison of Hellinger distance in Table $\ref{tab:hd}$, we would expect them to have similar performances. However, for OPT and BSP, their runtime would be a major bottleneck for their applicability  The situation for KDE is slightly more complicated: not only it is computationally quite intensive, its output can not be represented as concisely as partition based methods. Here we see that the efficiency of DSP makes it possible to utilize it for other machine learning tasks.

  \subsection{Mode Detection}
  A direct application of the piecewise constant density is to detect modes \cite{Comaniciu2002}, i.e., the dense areas or local maxima on the domain. The modes of our density estimator is defined as
  \begin{definition}
    A mode of the piecewise constant density is a sub-rectangle in the partition that its density is largest among all its neighbors as indicated by the stars in Figure \ref{simulation}.
    \label{mode_def}
  \end{definition}
  In order to compare our method with OPT and BSP in terms of performance in mode detection, we simulate samples from $x\sim (\sum_{i = 1}^4\pi_i \mathcal{N}_i(\mu_i, \Sigma))\mathbf{1}\{x\in [0, 1]^d\}$ with $d = \{2, 3, 4, 5, 6\}$ and $n = \{10^3, 10^4, 10^5\}$ respectively, where
  \begin{equation*}
  \left(\begin{array}{c}
  \mu_1\\
  \mu_2\\
  \mu_3\\
  \mu_4
  \end{array}\right) = \left(\begin{array}{rrrrr}
  1/4 & 1/4 & 1/2 & \cdots & 1/2\\
  1/4 & 3/4 & 1/2 & \cdots & 1/2\\
  3/4 & 1/4 & 1/2 & \cdots & 1/2\\
  3/4 & 3/4 & 1/2 & \cdots & 1/2\\
  \end{array}\right)_{4\times d}
  \end{equation*}
  and $\Sigma = 0.01\mathbf{I}$, where $\mathbf{I}$ is the identity matrix, $\pi = (1/4, 1/4, 1/4, 1/4)$. 
  The results of mode detection are summarized in Table \ref{tab1}, the standard error is obtained by generating 20 replicas for each $(d, n)$ pair. It is shown that density estimates by OPT and BSP have more modes generally. One possible explanation is that MAPs of OPT and BSP try to find the global optimizer of all possible binary partitions, they tend to overfit the data and result in a partition with many noisy sub-regions; in contrast, DSP makes myopic decisions and the possible choices for splitting are limited, but one can still bound the overall integration error by controlling the discrepancy adaptively as \eqref{cond}.
  \begin{table*}[ht]
  \centering
  \caption{The average number of modes detected by OPT, BSP and  DSP  for each pair $(d, n)$ respectively.}
  \label{tab1}
    \resizebox{\textwidth}{!}{
    
    \begin{tabular}{@{}cccccccccccc@{}}\toprule
  & \multicolumn{3}{c}{\#modes($n = 10^3$)} & \phantom{abc}& \multicolumn{3}{c}{\#modes($n = 10^4$)} &
  \phantom{abc} & \multicolumn{3}{c}{\#modes($n = 10^5$)}\\
 \midrule
  $d$ & OPT & BSP & DSP && OPT & BSP & DSP && OPT & BSP & DED\\\
  $2$ & 5.1(1.1) & 4.4(1.2) & 3.8(0.4) && 5.9(1.3) & 5.7(0.9) & 4.0(0.7) && 8.1(3.1) & 7.1(2.3) & 4.8(0.8)\\
  $3$ & 3.2(0.5) & 3.7(0.8) & 2.4(0.5) && 4.7(1.2) & 6.2(1.7) & 3.5(0.4) && 7.7(2.9) & 6.9(1.3) & 4.4(0.5)\\
  $4$ & 4.1(0.8) & 4.6(1.0) & 2.7(0.4) && 6.1(2.1) & 5.7(1.8) & 3.0(0.9) && 6.4(2.0) & 7.2(3.3) & 4.2(0.4)\\
  $5$ & 3.3(0.6) & 4.1(1.5) & 2.1(0.6) && 6.6(1.7) & 7.8(2.2) & 3.7(0.8) && 8.7(2.0) & 8.1(3.2) & 4.2(1.1)\\
  $6$ & 4.7(1.2) & 4.3(1.4) & 3.1(0.5) && 5.9(1.9) & 7.5(2.9) & 4.2(1.0) && 9.1(1.7) & 8.2(4.4) & 5.1(1.3)\\
  \bottomrule
  \end{tabular}
  }
  \end{table*}

  \begin{figure*}[ht]
    \center
    \includegraphics[width = 1.0\textwidth]{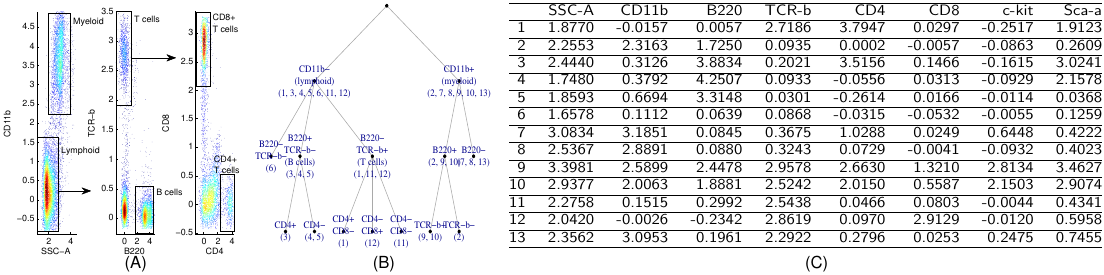}
    \caption{(A): an illustrative gating sequence, the cell type in each gate is attached; (B) there are 13 modes detected by our algorithm, and we arrange these modes into a hierarchical dendrogram: at first lvel, they are grouped by expression levels of CD11b; subsequently, the CD11b- modes are grouped according to B220 and TCR-b then further splitted according to CD4 and CD8 on the next level; the CD11b+ modes are grouped by B220 then by TCR-b; (C) the details of the expression levels of each mode.}
    \label{mouse}
  \end{figure*}
  \subsubsection{Flow Cytometry}
  Flow cytometry allows to measure simultaneously multiple characteristics of a large number of cells and is a ubiquitous and indispensable technology in medical settings. One effort of current research is to identify homogeneous sub-populations of cells automatically instead of manual gating, which is criticized for its subjectivity and non-scalability. There are a large amount of recent literatures concerning on auto gating and clustering, see \cite{Aghaeepour2013} and many references therein.

  In order to apply our method, we regard each cell as one observation in the sample space, i.e., if there are $n$ markers attached to a single cell, then the whole data set is generated from a hypothetical $n$ dimensional distribution. Mature cell populations concentrate in some high density areas, which can be easily identified in the binary partitioned space by Definition \ref{mode_def}.

  One practical issue needs to be addressed for most of the Cytometry analysis techniques: there is asymmetry in sub-populations; by optimizing a predefined loss function, it is possible that some sparse yet crucial populations are overlooked if the algorithms take most of the efforts to control the loss in denser areas. A remedy for this issue is to perform a down-sampling \cite{Aghaeepour2013, Qiu2011} step to roughly equalize the densities among populations then up-sampling after populations are identified; however, this step is dangerous that it may fails to sample enough cells in sparse populations, as a result, these populations are lost in the down-sampled data. In contrast, our approach does not require down-sampling step, and the asymmetry among populations are captured by their densities.

  For the mouse bone marrow data studied in \cite{Qiu2011}, we choose the 8 markers (SSA-C, CD11b, B220, TCR-$\beta$, CD4, CD8, c-kit, Sca-1) that are relevant to the cell types of interests; the number of cells is $\thicksim$380,000 after removing mutli-cell aggregates and co-incident events. 13 sub-populations are identified by our algorithm (\cite{Qiu2011} and its supplementary materials), the results are summarized in Figure \ref{mouse}.
  \subsubsection{Image Segmentation}
  Following \cite{Rubio2013} in which a new density estimation via histogram transforms is proposed, we conduct a similar experiment dealing with color image segmentation. The author in \cite{Rubio2013} reports that the result of his new algorithm is ``barely the same'' as that of others, thus we use mean shift with Gaussian kernel density estimator as the benchmark, which is publicly available in the GUI version of Edge Detection and Image SegmentatiON (EDISON) system \cite{Comaniciu2002}. For each pixel, we concatenate its LUV feature space representation with its coordinates to form a 5-dim \emph{joint domain} \cite{Comaniciu2002} representation. Our method are used to learn a 5-dim piecewise constant density. After identifying the modes according to Definition \ref{mode_def}, we use $k-$means to group the pixels with the metric
  \[d(x_1, x_2) = (\|x_1^r - x_2^r\|_2^2 + \lambda\|x_1^s - x_2^s\|_2^2)^{\frac{1}{2}}\]
  we write $x = (x^r, x^s)$ corresponding to the \emph{range} (color) domain and \emph{spatial} domain; $\lambda$ controls the relative importance of spatial difference, for example, a large $\lambda$ tends to connect adjacent pixels even if their colors are very different. Each cluster obtained from $k-$means corresponds to several patches in the original image; and each pixel is replaced by the average color in the patch it belongs to.

\begin{figure}[!ht]
 \centering
    \includegraphics[width = 0.50\textwidth]{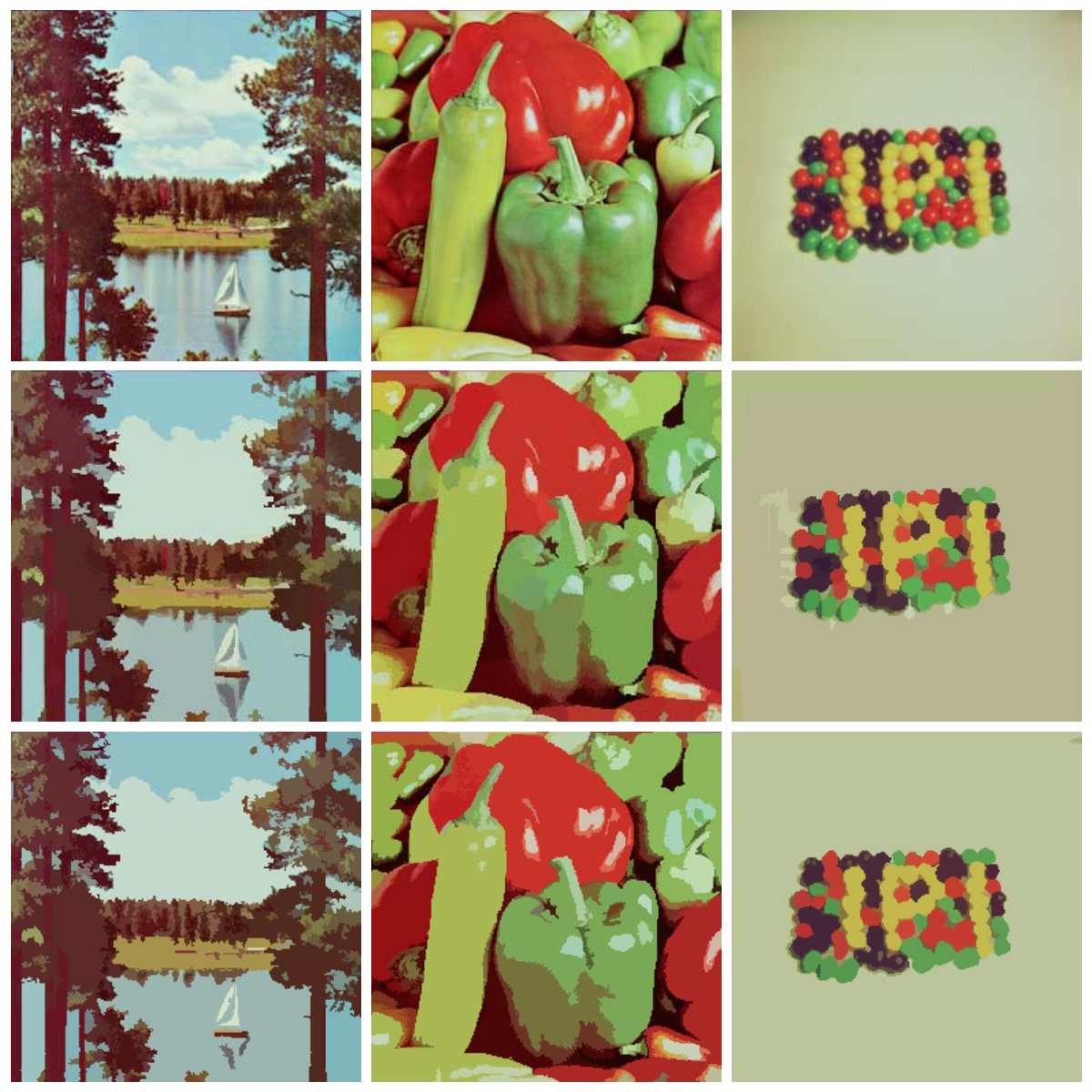}
    \caption{a) 1st row: original images; 2nd row: segmentation by mean shift with Gaussian kernel with default parameters; 3rd row: segmentation by DED. b) 1st column: lake; 2nd column: pepper; 3rd column: beans.}
    \label{imgseg}
  \end{figure}

  Once each pixel is process as above, some region connecting or pruning algorithms are employed to eliminate spurious patches. For easy of comparison, we employ the APIs in EDISON system to merge patches with its default parameters. The images are chosen from USC-SIPI Image Database \cite{Weber1997} and are rescaled to $256\times 256$ pixels by bicubic interpolation. The results are summarized in Figure \ref{imgseg}.

  \subsection{Some Other Applications}\label{soa}
  \textbf{Density Topology Exploration and Visualization.} The connectivity graph (DG) or level set tree \cite{Zhou2009} is widely used to represent energy landscapes of systems; it summarizes the hierarchy among various local maxima and minima in the configuration space; its topology is a tree and each inner node on the tree is a changing point that merges two or more independent regions in the domain. With the density estimation at hand, one may construct DG for samples instead of a given energy or density function. Unlike KDE that suffers from many local bumps and results in an overly complicated DG, $\eqref{eq1}$ is well suited for this purpose, partially because it smoothes out the minor fluctuations and takes only limited number of values; moreover, the simple structure of \eqref{eq1} makes the construction of such graph easy (i.e., one can just scan through each $r_i$ in decreasing order of $d(r_i)$). The DG of \eqref{eq1} not only reveals the modes of the density on its leaves, it also provides a tool to visualize high dimensional data hierarchically; for example, in fiber tractography \cite{Kent2013}, DG is used to visualize and analyze topography in fiber streamlines interactively.

\begin{figure}[!ht]
   \centering
    \includegraphics[width = 0.50\textwidth]{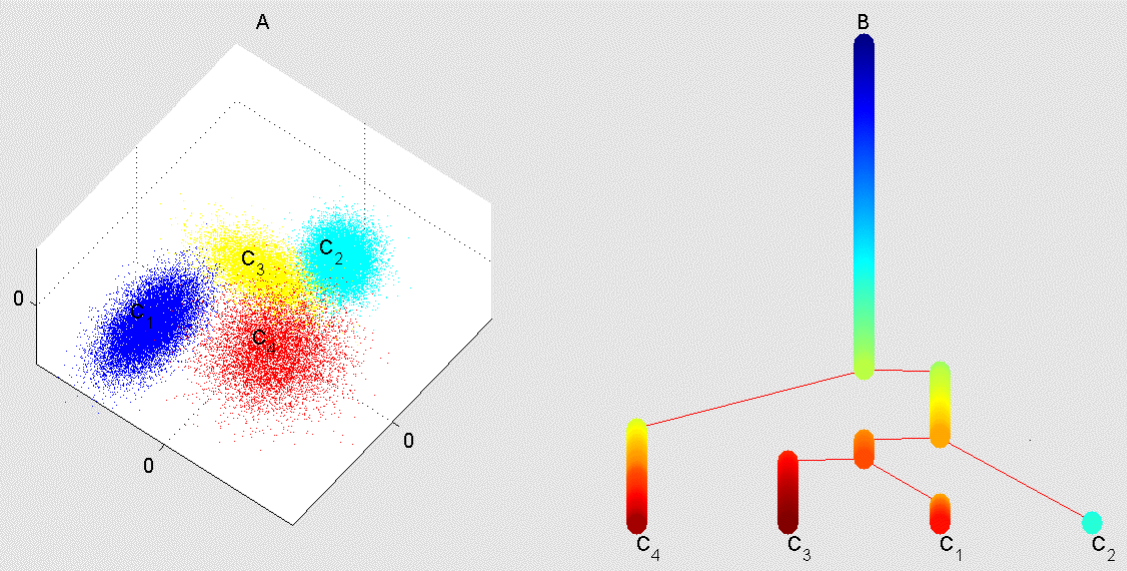}
    \caption{Left (A): the samples are generated from a Gaussian Mixture with 4 modes. Right (B): the level set tree. The clusters are annotated by $C_1, C_2, C_3, C_4$.}
    \label{DT}
  \end{figure}

  We demonstrate that how our piecewise density function can be used to construct level set trees in Figure \ref{DT}. The basic pipeline is to scan sub-rectangles sequentially according to the decreasing order of their densities and agglomerate the sub-rectangles according to their adjacency.

  \textbf{Multi-level Feature Extraction.} The density of each observation is available after learning the density function \eqref{eq1} and each sub-rectangle groups the observations with similar densities. These densities contains important non-linearity within the data which is hard to capture by standard transformations. We can augment the feature space of the sample by appending their corresponding densities. Through varying the deepest levels (Remark \ref{algo_rem_2}), the densities learned from different levels are included in the features; specifically, let $\hat{p}_{l_1}, ..., \hat{p}_{l_k}$ are learned densities of sample $\{x_i\}_{i = 1}^n$ by controlling the deepest levels to be $l_1, ..., l_k$ respectively, then the learned features are
  \[\{(\hat{p}_{l_1}(x_i), ..., \hat{p}_{l_k}(x_i))\}_{i = 1}^n\]
  This multi-level feature extraction technique has potential applications in representation learning.

\section{Conclusion}

In this paper we propose a novel density estimation method based on ideas from Quasi-Monte Carlo analysis. We prove it achieves a $O(n^{-\frac{1}{2}})$ error rate. By comparing it with other density estimation methods, we show DSP has comparable performance in terms of Hellinger distance while achieving a significant speed-up. We also show how DSP can be used to find good initializations for k-means. Due to space limitation, we were unable to include other interesting applications including mode seeking, data visualization via level set tree and data compression  \cite{gray1997vector}. 

\subsection*{Acknowledgements. { \normalsize\normalfont  This work was supported by NIH-R01GM109836, NSF-DMS1330132 and NSF-DMS1407557. Kun Yang's work was done when the author was a graduate student at Stanford University.}}

\newpage

\small
\renewcommand{\baselinestretch}{0.5}
\bibliography{dsp-arxiv}

\begin{thebibliography}{10}

\bibitem{Lu2013}
Luo Lu, Hui Jiang, and Wing~H Wong.
\newblock Multivariate density estimation by bayesian sequential partitioning.
\newblock {\em Journal of the American Statistical Association},
  108(504):1402--1410, 2013.

\bibitem{parzen1962estimation}
Emanuel Parzen.
\newblock On estimation of a probability density function and mode.
\newblock {\em The annals of mathematical statistics}, 33(3):1065--1076, 1962.

\bibitem{Wong2010}
Wing~H Wong and Li~Ma.
\newblock Optional p\'{o}lya tree and bayesian inference.
\newblock {\em The Annals of Statistics}, 38(3):1433--1459, 2010.

\bibitem{liu2011forest}
Han Liu, Min Xu, Haijie Gu, Anupam Gupta, John Lafferty, and Larry Wasserman.
\newblock Forest density estimation.
\newblock {\em The Journal of Machine Learning Research}, 12:907--951, 2011.

\bibitem{ram2011density}
Parikshit Ram and Alexander~G Gray.
\newblock Density estimation trees.
\newblock In {\em Proceedings of the 17th ACM SIGKDD international conference
  on Knowledge discovery and data mining}, pages 627--635. ACM, 2011.

\bibitem{Kuipers2012}
Lauwerens Kuipers and Harald Niederreiter.
\newblock {\em Uniform distribution of sequences}.
\newblock Courier Dover Publications, 2012.

\bibitem{Owen2005}
Art~B Owen.
\newblock Multidimensional variation for quasi-monte carlo.
\newblock In {\em International Conference on Statistics in honour of Professor
  Kai-Tai Fang's 65th birthday}, pages 49--74, 2005.

\bibitem{Heinrich2000}
Stefan Heinrich, Erich Novak, Grzegorz~W Wasilkowski, and Henryk Wozniakowski.
\newblock The inverse of the star-discrepancy depends linearly on the
  dimension.
\newblock {\em ACTA ARITHMETICA-WARSZAWA-}, 96(3):279--302, 2000.

\bibitem{Doerr2013}
Carola Doerr, Michael Gnewuch, and Magnus Wahlstr\'{o}m.
\newblock Calculation of discrepancy measures and applications.
\newblock {\em Preprint}, 2013.

\bibitem{Gnewuch2012a}
Michael Gnewuch.
\newblock Entropy, randomization, derandomization, and discrepancy.
\newblock In {\em Monte Carlo and quasi-Monte Carlo methods 2010}, pages
  43--78. Springer, 2012.

\bibitem{Liang2001}
Jia-Juan Liang, Kai-Tai Fang, Fred Hickernell, and Runze Li.
\newblock Testing multivariate uniformity and its applications.
\newblock {\em Mathematics of Computation}, 70(233):337--355, 2001.

\bibitem{Gnewuch2012}
Michael Gnewuch, Magnus Wahlstr\'{o}m, and Carola Winzen.
\newblock A new randomized algorithm to approximate the star discrepancy based
  on threshold accepting.
\newblock {\em SIAM Journal on Numerical Analysis}, 50(2):781--807, 2012.

\bibitem{arthur2007k}
David Arthur and Sergei Vassilvitskii.
\newblock k-means++: The advantages of careful seeding.
\newblock In {\em Proceedings of the eighteenth annual ACM-SIAM symposium on
  Discrete algorithms}, pages 1027--1035. Society for Industrial and Applied
  Mathematics, 2007.

\bibitem{katsavounidis1994new}
Ioannis Katsavounidis, C-C Jay~Kuo, and Zhen Zhang.
\newblock A new initialization technique for generalized lloyd iteration.
\newblock {\em Signal Processing Letters, IEEE}, 1(10):144--146, 1994.

\bibitem{fraley1998algorithms}
Chris Fraley.
\newblock Algorithms for model-based gaussian hierarchical clustering.
\newblock {\em SIAM Journal on Scientific Computing}, 20(1):270--281, 1998.

\bibitem{redmond2007method}
Stephen~J Redmond and Conor Heneghan.
\newblock A method for initialising the k-means clustering algorithm using
  kd-trees.
\newblock {\em Pattern recognition letters}, 28(8):965--973, 2007.

\bibitem{xu2015pca}
Qin Xu, Chris Ding, Jinpei Liu, and Bin Luo.
\newblock Pca-guided search for k-means.
\newblock {\em Pattern Recognition Letters}, 54:50--55, 2015.

\bibitem{su2007search}
Ting Su and Jennifer~G Dy.
\newblock In search of deterministic methods for initializing k-means and
  gaussian mixture clustering.
\newblock {\em Intelligent Data Analysis}, 11(4):319--338, 2007.

\bibitem{kaul2013building}
Manohar Kaul, Bin Yang, and Christian~S Jensen.
\newblock Building accurate 3d spatial networks to enable next generation
  intelligent transportation systems.
\newblock In {\em Mobile Data Management (MDM), 2013 IEEE 14th International
  Conference on}, volume~1, pages 137--146. IEEE, 2013.

\bibitem{aghaeepour2013critical}
Nima Aghaeepour, Greg Finak, Holger Hoos, Tim~R Mosmann, Ryan Brinkman, Raphael
  Gottardo, Richard~H Scheuermann, FlowCAP Consortium, DREAM Consortium, et~al.
\newblock Critical assessment of automated flow cytometry data analysis
  techniques.
\newblock {\em Nature methods}, 10(3):228--238, 2013.

\bibitem{spitzer2015interactive}
Matthew~H Spitzer, Pier~Federico Gherardini, Gabriela~K Fragiadakis, Nupur
  Bhattacharya, Robert~T Yuan, Andrew~N Hotson, Rachel Finck, Yaron Carmi,
  Eli~R Zunder, Wendy~J Fantl, et~al.
\newblock An interactive reference framework for modeling a dynamic immune
  system.
\newblock {\em Science}, 349(6244):1259425, 2015.

\bibitem{Comaniciu2002}
Dorin Comaniciu and Peter Meer.
\newblock Mean shift: A robust approach toward feature space analysis.
\newblock {\em Pattern Analysis and Machine Intelligence, IEEE Transactions
  on}, 24(5):603--619, 2002.

\bibitem{Aghaeepour2013}
Nima Aghaeepour, Greg Finak, Holger Hoos, Tim~R Mosmann, Ryan Brinkman, Raphael
  Gottardo, Richard~H Scheuermann, FlowCAP Consortium, and DREAM Consortium.
\newblock Critical assessment of automated flow cytometry data analysis
  techniques.
\newblock {\em Nature methods}, 2013.

\bibitem{Qiu2011}
Peng Qiu, Erin~F Simonds, Sean~C Bendall, Kenneth~D Gibbs~Jr, Robert~V
  Bruggner, Michael~D Linderman, Karen Sachs, Garry~P Nolan, and Sylvia~K
  Plevritis.
\newblock Extracting a cellular hierarchy from high-dimensional cytometry data
  with spade.
\newblock {\em Nature biotechnology}, 29(10):886--891, 2011.

\bibitem{Rubio2013}
E~L\'{o}pez-Rubio.
\newblock A histogram transform for probability density function estimation.
\newblock {\em IEEE transactions on pattern analysis and machine intelligence},
  2013.

\bibitem{Weber1997}
Allan Weber.
\newblock The usc-sipi image database.
\newblock {\em Signal and Image Processing Institute of the University of
  Southern California. URL: http://sipi. usc. edu/services/database}, 1997.

\bibitem{Zhou2009}
Qing Zhou and Wing~Hung Wong.
\newblock Energy landscape of a spin-glass model: Exploration and
  characterization.
\newblock {\em Physical Review E}, 79(5):051117, 2009.

\bibitem{Kent2013}
Brian~P Kent, Alessandro Rinaldo, Fang-Cheng Yeh, and Timothy Verstynen.
\newblock Mapping topographic structure in white matter pathways with level set
  trees.
\newblock {\em arXiv preprint arXiv:1311.5312}, 2013.

\bibitem{gray1997vector}
Robert~M Gray and Richard~A Olshen.
\newblock Vector quantization and density estimation.
\newblock In {\em Compression and Complexity of Sequences 1997. Proceedings},
  pages 172--193. IEEE, 1997.

\end{thebibliography}
\newpage

\end{document}